# Word Sense Disambiguation Using Conceptual Density

**Eneko Agirre\***
Lengoaia eta Sistema Informatikoak saila. Euskal Herriko Universitatea.
p.k. 649, 200800 Donostia. Spain. jibagbee@si.heu.es

**German Rigau\*\***
Departament de Llenguatges i Sistemes Informàtics. Universitat Politècnica de Catalunya.
Pau Gargallo 5, 08028 Barcelona. Spain. g.rigau@lsi.upc.es

## Abstract.

This paper presents a method for the resolution of lexical ambiguity of nouns and its automatic evaluation over the Brown Corpus. The method relies on the use of the wide-coverage noun taxonomy of WordNet and the notion of conceptual distance among concepts, captured by a Conceptual Density formula developed for this purpose. This fully automatic method requires no hand coding of lexical entries, hand tagging of text nor any kind of training process. The results of the experiments have been automatically evaluated against SemCor, the sense-tagged version of the Brown Corpus.

## 1 Introduction

Word sense disambiguation is a long-standing problem in Computational Linguistics. Much of recent work in lexical ambiguity resolution offers the prospect that a disambiguation system might be able to receive as input unrestricted text and tag each word with the most likely sense with fairly reasonable accuracy and efficiency. The most extended approach is to attempt to use the context of the word to be disambiguated together with information about each of its word senses to solve this problem.

Several interesting experiments have been performed in recent years using preexisting lexical knowledge resources: [Cowie et al. 92], [Wilks et al. 93] with LDOCE, [Yarowsky 92] with Roget's International Thesaurus, and [Sussna 93], [Voorhees 93], [Richarson et al. 94], [Resnik 95] with WordNet.

Although each of these techniques looks promising for disambiguation, either they have been only applied to a small number of words, a few sentences or not in a public domain corpus. For this reason we have tried to disambiguate all the nouns from real texts in the public domain sense tagged version of the Brown corpus [Francis & Kucera 67], [Miller et al. 93], also called Semantic Concordance or SemCor for short[1]. The words in SemCor are tagged with word senses from WordNet, a broad semantic taxonomy for English [Miller 90][2]. Thus SemCor provides an appropriate environment for testing our procedures and comparing among alternatives in a fully automatic way.

The automatic decision procedure for lexical ambiguity resolution presented in this paper is based on an elaboration of the conceptual distance among concepts: Conceptual Density [Agirre & Rigau 95]. The system needs to know how words are clustered in semantic classes, and how semantic classes are hierarchically organised. For this purpose, we have used WordNet. Our system tries to resolve the lexical ambiguity of nouns by finding the combination of senses from a set of contiguous nouns that maximises the total Conceptual Density among senses.

The performance of the procedure was tested on four texts from SemCor chosen at random. For comparison purposes two other approaches, [Sussna 93] and [Yarowsky 92], were also tried. The results show that our algorithm performs better on the test set.

Following this short introduction the Conceptual Density formula is presented. The main procedure to resolve lexical ambiguity of nouns using Conceptual Density is sketched on section 3. Section 4 describes extensively the experiments and its results. Finally, sections 5 and 6 deal with further work and conclusions.

---

[1] Semcor comprises approximately 250,000 words. The tagging was done manually, and the error rate measured by the authors is around 10% for polysemous words.
[2] The senses of a word are represented by synsets, one for each word sense. The nominal part of WordNet can be viewed as a tangled hierarchy of hypo/hypernymy relations. Nominal relations include also three kinds of meronymic relations, which can be paraphrased as member-of, made-of and component-part-of. The version used in this work is WordNet 1.4, The coverage in WordNet of the senses for open-class words in SemCor reaches 96% according to the authors.

## 2 Conceptual Density and Word Sense Disambiguation

A measure of the relatedness among concepts can be a valuable prediction knowledge source for several decisions in Natural Language Processing. For example, the relatedness of a certain word-sense to the context allows us to select that sense over the others, and actually disambiguate the word. As was pointed by [Miller & Teibel, 91], relatedness can be measured by a fine-grained conceptual distance among concepts in a hierarchical semantic net such as WordNet. This measure would allow to discover reliably the lexical cohesion of a given set of words in English.

Conceptual distance tries to provide a basis for determining closeness in meaning among pairs of words, taking as reference a structured hierarchical net. Conceptual distance between two concepts is defined in [Rada et al. 89] as the length of the shortest path that connects the concepts in a hierarchical semantic net. In a similar approach, [Sussna 93] employs the notion of conceptual distance between network nodes in order to improve precision during document indexing. [Resnik 95] captures semantic similarity (closely related to conceptual distance) by means of the information content of the concepts in a hierarchical net. In general these approaches focus on nouns.

The measure of conceptual distance among concepts we are looking for should be sensitive to:

• the length of the shortest path that connects the concepts involved.
• the depth in the hierarchy: concepts in a deeper part of the hierarchy should be ranked closer.
• the density of concepts in the hierarchy: concepts in a dense part of the hierarchy are relatively closer than those in a more sparse region.
• the measure should be independent of the number of concepts we are measuring.

We have experimented with several formulas that follow the four criteria presented above. The experiments reported here were performed using the Conceptual Density formula [Agirre & Rigau 95], which compares areas of subhierarchies.

To illustrate how Conceptual Density can help to disambiguate a word, in figure 1 the word W has four senses and several context words. Each sense of the words belongs to a subhierachy of WordNet. The dots in the subhierarchies represent the senses of either the word to be disambiguated (W) or the words in the context. Conceptual Density will yield the highest density for the subhierarchy containing more senses of those, relative to the total amount of senses in the subhierarchy. The sense of W contained in the subhierarchy with highest Conceptual Density will be chosen as the sense disambiguating W in the given context. In figure 1, sense2 would be chosen.

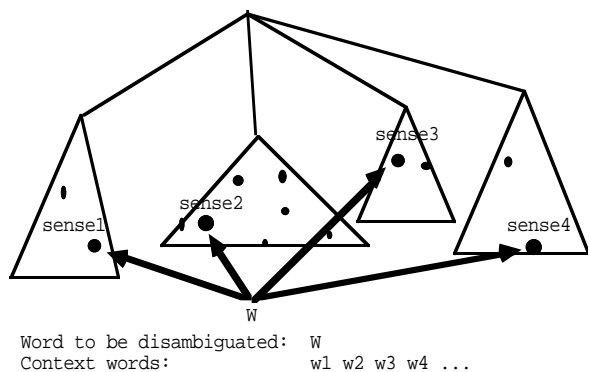

Word to be disambiguated:  W
Context words:          w1 w2 w3 w4 ...

Figure 1: senses of a word in WordNet

Given a concept $c$, at the top of a subhierarchy, and given $nhyp$ and $h$ (mean number of hyponyms per node and height of the subhierarchy, respectively), the Conceptual Density for $c$ when its subhierarchy contains a number $m$ (marks) of senses of the words to disambiguate is given by the formula below:

$$CD(c,m) = \frac{\sum_{i=0}^{m-1} nhyp^{i^{0.20}}}{descendants_c} \qquad (1)$$

Formula 1 shows a parameter that was computed experimentally. The 0.20 tries to smooth the exponential, as $m$ ranges between 1 and the total number of senses in WordNet. Several values were tried for the parameter, and it was found that the best performance was attained consistently when the parameter was near 0.20.

## 3 The Disambiguation Algorithm Using Conceptual Density

Given a window size, the program moves the window one noun at a time from the beginning of the document towards its end, disambiguating in each step the noun in the middle of the window and considering the other nouns in the window as context. Non-noun words are not taken into account.

```
(Step 1) tree := compute_tree(words_in_window)
         loop
(Step 2)   tree := compute_conceptual_distance(tree)
(Step 3)   concept := selecct_concept_with_highest_weigth(tree)
           if  concept = null then exitloop
(Step 4)   tree := mark_disambiguated_senses(tree,concept)
         endloop
(Step 5) output_disambiguation_result(tree)
```

Figure 2: algorithm for each window

The algorithm to disambiguate a given noun w in the middle of a window of nouns W (c.f. figure 2) roughly proceeds as follows. First, the algorithm represents in a lattice the nouns present in the window, their senses and hypernyms (step 1). Then, the program computes the Conceptual Density of each concept in WordNet according to the senses it contains in its subhierarchy (step 2). It selects the concept c with highest Conceptual Density (step 3) and selects the senses below it as the correct senses for the respective words (step 4).

The algorithm proceeds then to compute the density for the remaining senses in the lattice, and continues to disambiguate the nouns left in W (back to steps 2, 3 and 4). When no further disambiguation is possible, the senses left for w are processed and the result is presented (step 5).

Besides completely disambiguating a word or failing to do so, in some cases the disambiguation algorithm returns several possible senses for a word. In the experiments we considered these partial outcomes as failure to disambiguate.

## 4  The Experiments

### 4.1  The texts

We selected four texts from SemCor at random: br-a01 (where a stands for the gender "Press: Reportage"), br-b20 (b for "Press: Editorial"), br-j09 (j means "Learned: Science") and br-r05 (r for "Humour"). Table 1 shows some statistics for each text

| text | words | nouns | nouns in WN | monosemous |
|------|-------|-------|-------------|------------|
| br-a01 | 2079 | 564 | 464 | 149 (32%) |
| br-b20 | 2153 | 453 | 377 | 128 (34%) |
| br-j09 | 2495 | 620 | 586 | 205 (34%) |
| br-r05 | 2407 | 457 | 431 | 120 (27%) |
| total | 9134 | 2094 | 1858 | 602 (32%) |

Table 1: data for each text

An average of 11% of all the nouns in these four texts were not found in WordNet. According to this data, the amount of monosemous nouns in these texts is bigger (32% average) than the one calculated for the open-class words from the whole SemCor (27.2% according to [Miller et al. 94]).

For our experiments, these texts play both the role of input files (without semantic tags) and (tagged) test files. When they are treated as input files, we throw away all non-noun words, only leaving the lemmas of the nouns present in WordNet.

```
The jury praised the administration and operation of the Atlanta
Police_Department, the Fulton_Tax_Commissioner_'s_Office, the
Bellwood and Alpharetta prison_farms, Grady_Hospital and the
Fulton_Health_Department.
```

Figure 3: sample sentence from SemCor

```
<s>
<wd>jury</wd><sn>[noun.group.0]</sn><tag>NN</tag>
<wd>administration</wd><sn>[noun.act.0]</sn><tag>NN</tag>
<wd>operation</wd><sn>[noun.state.0]</sn><tag>NN</tag>
<wd>Police_Department</wd><sn>[noun.group.0]</sn><tag>NN</tag>
<wd>prison_farms</wd><mwd>prison_farm</mwd><msn>[noun.artifact.0]</msn><tag>NN</tag>
</s>
```

Figure 4: SemCor format

```
jury administration operation Police_Department prison_farm
```

Figure 5: input words

Figure 4 shows the SemCor format for the nouns in the example sentence in figure 3. The result of erasing irrelevant information obtaining the words[3] as they will be input to the algorithm are shown in figure 5.The output of the algorithm comprises sense tags that can be compared automatically with the original file (c.f. figure 4).

## 4.2 Results and evaluation

One of the goals of the experiments was to decide among different variants of the Conceptual Density formula. Results are given averaging the results of the four files. Partial disambiguation is treated as failure to disambiguate. Precision (that is, the percentage of actual answers which were correct) and recall (that is, the percentage of possible answers which were correct) are given in terms of polysemous nouns only. The graphs are drawn against the size of the context[4] that was taken into account when disambiguating.

• **meronymy does not improve performance as expected.** One parameter controls whether meronymic relations, in addition to the hypo/hypernymy relation, are taken into account or not. A priori, the more relations are taken in account the better density would capture semantic relatedness, and therefore better results can be expected. The experiments, see figure 6, showed that there is not much difference; adding meronymic information does not improve precision, and raises coverage only 3% (approximately). Nevertheless, in the rest of the results reported below, meronymy and hypernymy were used.

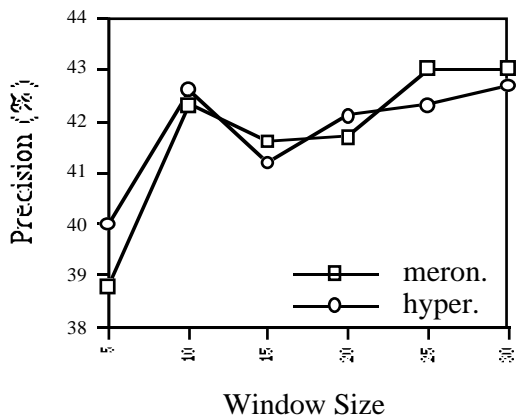

Figure 6: meronymy and hyperonymy

• **global nhyp is as good as local nhyp.** There was an aspect of the density formula which we could not decide analytically, and which we wanted to check experimentally. The average number of hyponyms or *nhyp* (c.f. formula 1) can be approximated in two ways. If an independent *nhyp* is computed for every concept in WordNet we call it *local nhyp*. If instead, a unique *nhyp* is computed using the whole hierarchy, we have *global nhyp*. While *local nhyp* is the actual average for a given concept, *global nhyp* gives only an estimation. The results (c.f. figure 7) show that *local nhyp* performs only slightly better. Therefore *global nhyp* is favoured and was used in subsequent experiments.

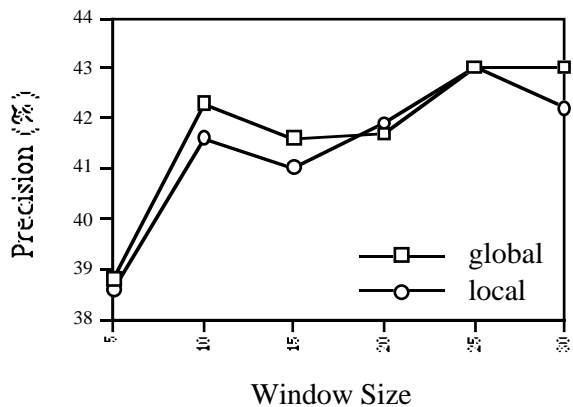

Figure 7: *local nhyp* vs. *global nhyp*

• **context size: different behaviour for each text.** Deciding the optimum context-size for disambiguating using Conceptual Density is an important issue. One could assume that the more context there is, the better the disambiguation results would be. Our experiments show that each file from SemCor has a different behaviour (c.f. figure 8) while br-b20 shows clear improvement for bigger window sizes, br-r05 gets a local maximum at a 10 size window, etc.

As each text is structured a list of sentences, lacking any indication of headings, sections, paragraph endings, text changes, etc. the program gathers the context without knowing whether the nouns actually occur in coherent pieces of text. This could account for the fact that in br-r05, composed mainly by short pieces of dialogues, the best results are for window size 10, the average size of this dialogue pieces. Longer windows will include other pieces of unrelated dialogues that could mislead the disambiguation.

Besides, the files can be composed of different pieces of unrelated texts without pointing it explicitly. For instance, two of our test files (br-a01 and br-b20) are collections of short journalistic texts.

This could explain that the performance of br-a01 decreases for windows of 30 nouns, because for most of the nouns the context would include nouns from another article.

The polysemy level could also affect the performance, but in our texts less polysemy does not correlate with better performance. Nevertheless the actual nature of each text is for sure an important factor, difficult to measure, which could account for the different behaviour on its own. For instance, the poor performance on text br-j09 could be explained by its technical nature. Further analysis of the errors, contexts and relations found among the words would be needed to be more conclusive.

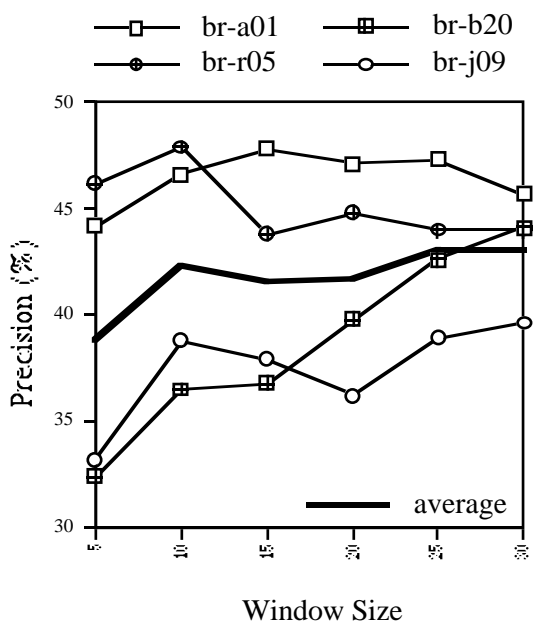

Figure 8: context size and different files

Leaving aside these considerations, and in order to give an overall view of the performance, we consider the average behaviour in order to lead our conclusions.

• **file vs. sense.** WordNet groups senses in 24 lexicographer's files. The algorithm assigns a noun both an specific sense and a file label. Both file matches and sense matches are interesting to count. While the sense level gives a fine graded measure of the algorithm, the file level gives an indication of the performance if we were interested in a less sharp level of disambiguation. The granularity of the sense distinctions made in [Hearst, 91], [Gale et al. 93] and [Yarowsky 92], also called homographs in [Guthrie et al. 93], can be compared to that of the file level in WordNet.

For instance, in [Yarowsky 92] two homographs of the noun *bass* are considered, one characterised as MUSIC and the other as ANIMAL, INSECT. In WordNet, the 6 senses of *bass* related to music appear in the following files: ARTIFACT, ATTRIBUTE, COMMUNICATION and PERSON. The 3 senses related to animals appear in the files ANIMAL and FOOD. This means that while the homograph level in [Yarowsky 92] distinguishes two sets of senses, the file level in WordNet distinguishes six sets of senses, still finer in granularity.

The following figure shows that, as expected, file-level matches attain better performance (71.2% overall and 53.9% for polysemic nouns) than sense-level matches.

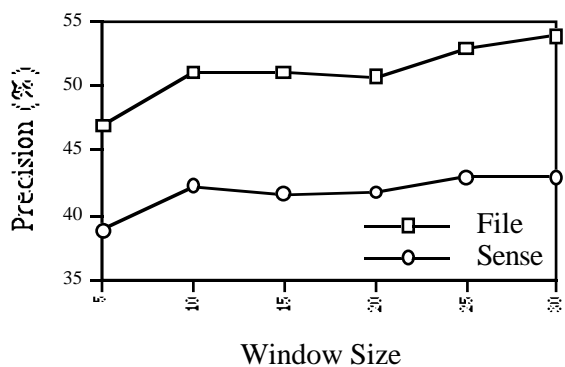

Figure 9: sense level vs. file level

• **evaluation of the results**

Figure 10 shows that, overall, coverage over polysemous nouns increases significantly with the window size, without losing precision. Coverage tends to get stabilised near 80%, getting little improvement for window sizes bigger than 20.

The figure also shows the guessing baseline, given by selecting senses at random. This baseline was first calculated analytically and later checked experimentally. We also compare the performance of our algorithm with that of the "most frequent" heuristic. The frequency counts for each sense were collected using the rest of SemCor, and then applied to the four texts. While the precision is similar to that of our algorithm, the coverage is 8% worse.

All the data for the best window size can be seen in table 2. The precision and coverage shown in all the preceding graphs were relative to the polysemous nouns only. If we also include monosemic nouns precision raises, as shown in table 2, from 43% to 64.5%, and the coverage increases from 79.6% to 86.2%.

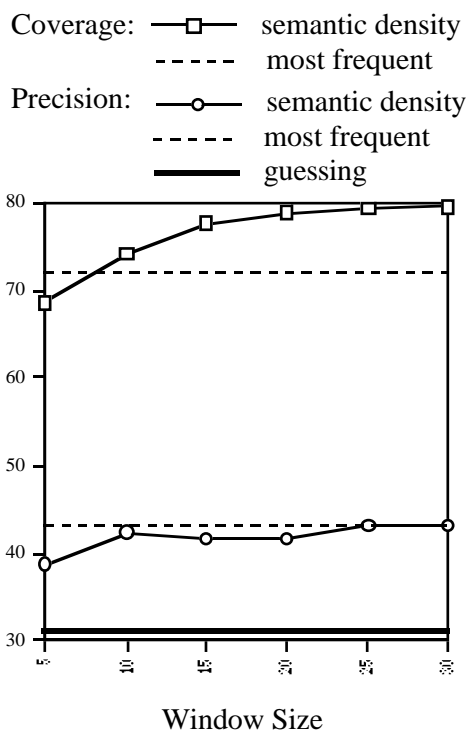

Coverage: ——□—— semantic density
- - - - - most frequent

Precision: ——○—— semantic density
- - - - - most frequent
——— guessing

**Window Size**

Figure 10: precision and coverage

| % | w=30 | | Cover. | Prec. | Recall |
|---|---|---|---|---|---|
| overall | File | | 86.2 | 71.2 | 61.4 |
| | Sense | | | 64.5 | 55.5 |
| polysemic | File | | 79.6 | 53.9 | 42.8 |
| | Sense | | | 43 | 34.2 |

Table 2: overall data for the best window size

## 4.3 Comparison with other works

The raw results presented here seem to be poor when compared to those shown in [Hearst 91], [Gale et al. 93] and [Yarowsky 92]. We think that several factors make the comparison difficult. Most of those works focus in a selected set of a few words, generally with a couple of senses of very different meaning (coarse-grained distinctions), and for which their algorithm could gather enough evidence. On the contrary, we tested our method with **all** the nouns in a subset of an unrestricted public domain corpus (more than 9.000 words), making fine-grained distinctions among all the senses in WordNet.

An approach that uses hierarchical knowledge is that of [Resnik 95], which additionally uses the information content of each concept gathered from corpora. Unfortunately he applies his method on a different task, that of disambiguating sets of related nouns. The evaluation is done on a set of related nouns from Roget's Thesaurus tagged by hand. The fact that some senses were discarded because the human judged them not reliable makes comparison even more difficult.

In order to compare our approach we decided to implement [Yarowsky 92] and [Sussna 93], and test them on our texts. For [Yarowsky 92] we had to adapt it to work with WordNet. His method relies on cooccurrence data gathered on Roget's Thesaurus semantic categories. Instead, on our experiment we use saliency values[5] based on the lexicographic file tags in SemCor (c.f. figure 4). The results for a window size of 50 are those shown in table 3[6]. The precision attained by our algorithm is higher. To compare figures better consider the results in table 4, were the coverage of our algorithm was easily extended using the version presented below, increasing recall to 70.1%.

| % | | Cover. | Prec. | Recall |
|---|---|---|---|---|
| C.Density | | 86.2 | 71.2 | 61.4 |
| Yarowsky | | 100.0 | 64.0 | 64.0 |

Table 3: comparison with [Yarowsky 92]

From the methods based on Conceptual Distance, [Sussna 93] is the most similar to ours. Sussna disambiguates several documents from a public corpus using WordNet. The test set was tagged by hand, allowing more than one correct senses for a single word. The method he uses has to overcome a combinatorial explosion[7] controlling the size of the window and "freezing" the senses for all the nouns preceding the noun to be disambiguated. In order to freeze the winning sense Sussna's algorithm is forced to make a unique choice. When Conceptual Distance is not able to choose a single sense, he has to choose one at random.

Conceptual Density overcomes the combinatorial explosion extending the notion of conceptual distance from a pair of words to n words, and therefore can yield more than one correct sense for a word. For comparison, we altered our algorithm to also make random choices when unable to choose a single sense. We applied the algorithm Sussna considers best, discarding the factors that do not affect performance significantly[8], and obtain the results in table 4.

---

[5] We tried both mutual information and association ratio, and the later performed better.

[6] The results of our algorithm are those for window size 30, file matches and overall.

[7] In our replication of his experiment the mutual constraint for the first 10 nouns (the optimal window size according to his experiments) of file br-r05 had to deal with more than 200.000 synset pairs.

[8] Initial mutual constraint size is 10 and window size is 41. Meronymic links are also considered. All the links have the same weigth.

| % | | Cover. | Prec. |
|---|---|---|---|
| C.Density | File | 100.0 | 70.1 |
| | Sense | | 60.1 |
| Sussna | File | 100.0 | 64.5 |
| | Sense | | 52.3 |

Table 4: comparison with [Sussna 93]

A more thorough comparison with these methods could be desirable, but not possible in this paper for the sake of conciseness.

# 5 Further Work

We would like to have included in this paper a study on whether there is or not a correlation among correct and erroneous sense assignations and the degree of Conceptual Density, that is, the actual figure held by formula 1. If this was the case, the error rate could be further decreased setting a certain threshold for Conceptual Density values of winning senses. We would also like to evaluate the usefulness of partial disambiguation: decrease of ambiguity, number of times correct sense is among the chosen ones, etc.

There are some factors that could raise the performance of our algorithm:

• **Work on coherent chunks of text.** Unfortunately any information about discourse structure is absent in SemCor, apart from sentence endings. If coherent pieces of discourse were taken as input, both performance and efficiency of the algorithm might improve. The performance would gain from the fact that sentences from unrelated topics would not be considered in the disambiguation window. We think that efficiency could also be improved if the algorithm worked on entire coherent chunks instead of one word at a time.

• **Extend and improve the semantic data.** WordNet provides sinonimy, hypernymy and meronymy relations for nouns, but other relations are missing. For instance, WordNet lacks cross-categorial semantic relations, which could be very useful to extend the notion of Conceptual Density of nouns to Conceptual Density of words. Apart from extending the disambiguation to verbs, adjectives and adverbs, cross-categorial relations would allow to capture better the relations among senses and provide firmer grounds for disambiguating.

These other relations could be extracted from other knowledge sources, both corpus-based or MRD-based, such as topic information (as can be found in Roget's Thesaurus), word frequencies, collocations [Yarowsky 93], selectional restrictions [Ribas 95], etc. If those relations could be given on WordNet senses, Conceptual Density could profit from them. [Richardson et al. 94] tries to combine WordNet and informational measures taken from corpora, defining a conceptual similarity considering both, but does not give any evaluation of their method. It is our belief, following the ideas of [McRoy 92] that full-fledged lexical ambiguity resolution should combine several information sources. Conceptual Density might be only one of a number of complementary evidences of the plausibility of a certain word sense.

• **Tune the sense distinctions to the level best suited for the application.** On the one hand the sense distinctions made by WordNet 1.4 are not always satisfactory and obviously, WordNet 1.4 is not a complete lexical database. On the other hand, our algorithm is not designed to work on the file level, e.g. if the sense level is unable to distinguish among two senses, the file level also fails, even if both senses were from the same file. If the senses were collapsed at the file level, the coverage and precision of the algorithm at the file level might be even better.

# 6 Conclusion

The automatic method for the disambiguation of nouns presented in this paper is ready-usable in any general domain and on free-running text, given part of speech tags. It does not need any training and uses word sense tags from WordNet, an extensively used lexical data base. The algorithm is theoretically motivated and founded, and offers a general measure of the semantic relatedness for any number of nouns.

Conceptual Density has been used for other tasks apart from the disambiguation of free-running test. Its application for automatic spelling correction is outlined in [Agirre et al. 94]. It was also used on Computational Lexicography, enriching dictionary senses with semantic tags extracted from WordNet [Rigau 94], or linking bilingual dicitonaries to WordNet [Rigau and Agirre 95]

In the experiments, the algorithm disambiguated four texts (more than 10.000 words long) of SemCor, a subset of the Brown corpus. The results were obtained automatically comparing the tags in SemCor with those computed by the algorithm, which would allow the comparison with other disambiguation methods. Two other methods, [Sussna 93] and [Yarowsky 92], were also tried on the same texts, showing that our algorithm performs better.

The results are promising, considering the difficulty of the task (free running text, large number of senses per word in WordNet), and the lack of any discourse structure of the texts. Two kinds of results can be obtained: the specific sense or a coarser, file level, tag.

## Acknowledgements

We wish to thank all the staff of the Computer Research Laboratory and specially Jim Cowie, Joe Guthrie, Louise Guthrie and David Farwell. We would also like to thank Xabier Arregi, Jose mari Arriola, Xabier Artola, Arantza Díaz de Ilarraza, Kepa Sarasola and Aitor Soroa from the Computer Science Faculty of EHU and Francesc Ribas, Horacio Rodríguez and Alicia Ageno from the Computer Science Department of UPC.